\documentclass{article} 
\usepackage[preprint]{colm2025_conference}

\usepackage{microtype}
\usepackage{hyperref}
\usepackage{url}
\usepackage{booktabs}

\usepackage{graphicx}
\usepackage{amsmath}

\usepackage{amssymb}
\usepackage{booktabs}

\usepackage{lineno}
\usepackage{float}

\usepackage{caption}
\usepackage{wrapfig}

\definecolor{darkblue}{rgb}{0, 0, 0.5}
\hypersetup{colorlinks=true, citecolor=darkblue, linkcolor=darkblue, urlcolor=darkblue}

\title{Focus Directions Make Your Language Models Pay More Attention to Relevant Contexts}

\author{Youxiang Zhu\textsuperscript{1}, Ruochen Li\textsuperscript{2}, Danqing Wang\textsuperscript{3}, Daniel Haehn\textsuperscript{1}, Xiaohui Liang\textsuperscript{1} \\
\textsuperscript{1}University of Massachusetts Boston, \textsuperscript{2}Technische Universität München,\\ \textsuperscript{3}Carnegie Mellon University\\
}

\begin{document}

\ifcolmsubmission
\linenumbers
\fi

\maketitle

\begin{abstract}
Long-context large language models (LLMs) are prone to be distracted by irrelevant contexts. The reason for distraction remains poorly understood. In this paper, we first identify the contextual heads, a special group of attention heads that control the overall attention of the LLM. Then, we demonstrate that distraction arises when contextual heads fail to allocate sufficient attention to relevant contexts and can be mitigated by increasing attention to these contexts.
We further identify focus directions, located at the key and query activations of these heads, which enable them to allocate more attention to relevant contexts without explicitly specifying which context is relevant. 
We comprehensively evaluate the effect of focus direction on various long-context tasks and find out focus directions could help to mitigate the poor task alignment of the long-context LLMs.
We believe our findings could promote further research on long-context LLM alignment.

\end{abstract}

\section{Introduction}

Long-context large language models enable multiple applications, such as many-shot in-context learning ~\cite{li2024long, agarwal2025many, bertsch2024context}, summarization ~\cite{chang2023booookscore,kim2024fables}, and retrieval-augmented generation ~\cite{lee2024can}. Given a long context window such as 128k tokens, only a small amount of the contexts are relevant to the task, and a large amount of contexts are irrelevant. Long context LLM may be distracted by irrelevant contexts ~\cite{liu2024lost, shi2023large}. Such distraction may result in generating false information, error reasoning, and negative social impacts.

\begin{figure}[]
  \centering
  \includegraphics[width=0.80\textwidth]{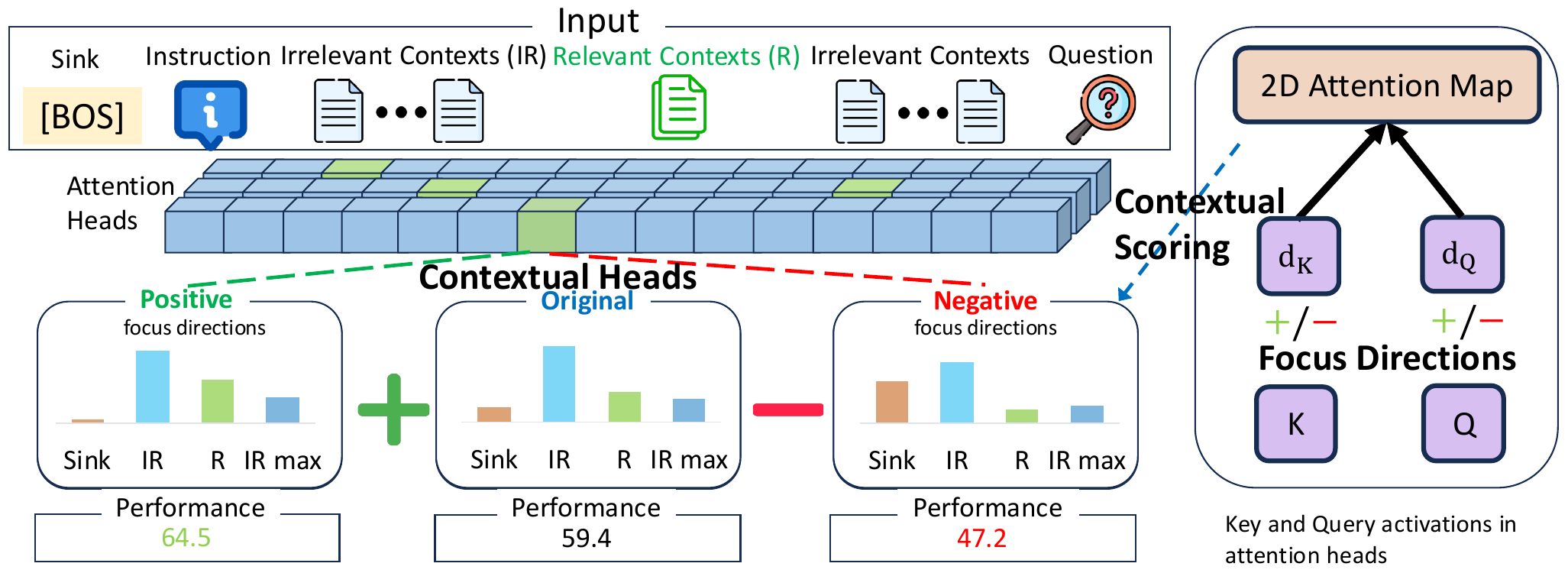}
  \caption{Overview of this work. We first introduce \textbf{contextual scoring}, measuring the attention distribution over inputs during response generation. Based on contextual scoring, we identify the \textbf{contextual heads}, which control the overall attention of LLMs. We further find out \textbf{focus directions}, which make LLMs pay more attention to the relevant contexts.}
  \label{fig:intro}
\end{figure}

The reason for LLMs being distracted by irrelevant context is poorly understood. In this paper, we aim to reveal the \textbf{cause of the distraction} (\S\ref{sec:casue_of_distraction}). As shown in Figure \ref{fig:intro}, starting with a dataset with labels of relevant and irrelevant context, we first introduce a contextual scoring method, which measures the strength of the attention to the relevant context during text generation. Based on such a scoring method, we identify \textbf{contextual heads}, a special group of attention heads with the highest score. We then adjust the strength of attention of these heads to the relevant contexts based on the label of the relevant context. We found that increasing attention on these heads to the relevant context increases the downstream task performance, while decreasing attention decreases the performance. While other non-contextual heads have minimal such effects. We conclude that contextual heads could control the overall attention of LLMs to the contexts.

Building upon the finding of the contextual head, we further wonder if a mechanism exists that could make LLMs figure out which is the relevant context by itself instead of relying on the external labels.
Recent works on activation addition show that LLMs' behavior, such as refusal ~\cite{arditi2024refusal}, sentiment ~\cite{han2023word}, and truthfulness ~\cite{li2024inference} can be changed by adding or subtracting a single directional vector in some intermediate activation space, following the linear representation hypothesis ~\cite{park2023linear}. 
Inspired by such works, we hypothesize and verify the existence of \textbf{focus directions} (\S\ref{sec:focus_direction}), which could enable LLMs to pay more attention to the relevant contexts.
Focus directions are located at the key and query activations of the transformer attention heads. We found that adding focus directions vector to the key and query activation could increase the attention to the relevant context, and subtracting a direction could decrease the attention to the relevant context. 
Such focus directions enable the LLMs' attention behavior control at inference time.

To understand how focus directions affect the capability of long-context LLMs (\S\ref{sec:focus_direction_tasks}), we apply focus directions to three families of LLMs and evaluate them on HELMET ~\cite{yen2024helmet}, a comprehensive long-context task benchmark. We found that focus directions could help mitigate poor task alignment of the LLMs. At last, we discuss the potential application of the focus directions.

\section{Cause of distraction}
\label{sec:casue_of_distraction}

To reveal the cause of LLMs being distracted by irrelevant contexts, we first identify the attention heads that are mostly responsible for extracting information from relevant contexts, which we named contextual heads \S\ref{sec:identify_contextual_heads}. Then, we study the basic properties of the contextual heads, including their location and behavior in different cases \S\ref{sec:properties_of_contextual_heads}. At last, we demonstrate that increasing attention to relevant contexts on these heads could mitigate distractions \S\ref{sec:attention_compensation}. 

\subsection{Identifying contextual heads}
\label{sec:identify_contextual_heads}

To identify contextual heads, we introduce a contextual scoring method to identify the attention distribution of different parts of input for each attention head in the transformer architecture. Our method is based on the Multi-Document Question Answering (QA) data introduced by the ``lost in the middle'' paper ~\cite{liu2024lost}.

\textbf{Multi-Document Question Answering data.} 
The data is initiated with the NaturalQuestions-Open data~\cite{lee2019latent, kwiatkowski2019natural}. Each samples have a question and a list of answers. The questions are user queries from Google search, and the answers are human-annotated based on Wikipedia. The authors of ~\cite{liu2024lost} further matched each question and answer pair with a set of documents using a retrieval system. In these documents, only one contains the answer (i.e., relevant context), and others do not contain the answer (i.e., irrelevant contexts). 

\textbf{Experiment settings.}
The above dataset has 2654 samples in total, we randomly split them into half training and half testing. 
The input is defined as $[I_p, \hat{C}_{before}, C, \hat{C}_{after}, I_q]$, where $I_p$ and $I_q$ are instructions, specifying the QA task (e.g., a system prompt and a question). The $C$ stands for relevant context, $\hat{C}_{before}$, and $\hat{C}_{after}$ stands for the irrelevant contexts before and after the relevant context, which can be zero, one or more documents. We consider 20 document cases where one of the documents in the input is relevant and the rest of the 19 documents are irrelevant. We put the relevant documents in positions 1, 5, 10, 15, and 20. The input is fed into an LLM, in our case we use Llama-3.2-3B instruction model\footnote{https://huggingface.co/meta-llama/Llama-3.2-3B-Instruct}, to obtain an LLM response $R$ using greedy decoding. 
The evaluation metric is the exact match (EM) accuracy. If the model output matches one of the answers in the output list, then it is considered to be correct; otherwise, it is wrong.

\textbf{Contextual scoring.}
Based on the above data and experiment settings, we introduce the following contextual scoring method, which aims to find a set of attention heads in the LLM that pay the most attention to the relevant contexts during generation. 
Let \( W \in \mathbb{R}^{T \times T} \) be the attention weight matrix of an attention head, where \( T \) is the sequence length. For each token \( r_i \) in the generated response \( R = [r_{start}, \dots, r_{end}] \), we extract the attention weights corresponding to the relevant context \( C = [c_{start}, \dots, c_{end}]\) and sum over this span, and then average through each response token $r_i$:
\begin{equation}
S_C = \frac{1}{|R|} \sum_{i=r_{start}}^{r_{end}} {\sum_{j=c_{start}}^{c_{end}} {W_{i,j}}}
\label{eq:relevant_score}
\end{equation}
This score quantifies how much an attention head focuses on the relevant context while generating the response. Higher values indicate stronger attention toward the relevant span, helping to identify heads that extract the most information from the relevant contexts. We then further average the score $S_C$ through the dataset for each head, obtaining a \textbf{relevant contextual score}. We do not normalize the score by length since, at the dataset level, each document does not have a significant difference in length. With such a score, we are now able to identify the \textbf{contextual heads} with top-$k$ scores focused on the relevant contexts. Similarly, we can extend the definition of the relevant contextual score to any text span in the input.
We could define \textbf{irrelevant contextual score}, which measures the attention to the entire irrelevant contexts (i.e., $\hat{C}_{before}$ and $\hat{C}_{after}$); \textbf{max single document irrelevant contextual score}, which represents the highest contextual score among individual documents within the irrelevant contexts; \textbf{sink contextual score}, which measure the ``dummy'' attention to the attention sink (i.e., starts tokens) ~\cite{xiao2023efficient} when that part of attention do not need to pay in other non-start tokens.

\subsection{Properties of contextual heads}
\label{sec:properties_of_contextual_heads}

\begin{wrapfigure}{r}{0.5\textwidth}
\vspace{-4mm}
\centering
\begin{minipage}{0.5\textwidth}
\centering
\includegraphics[width=\linewidth]{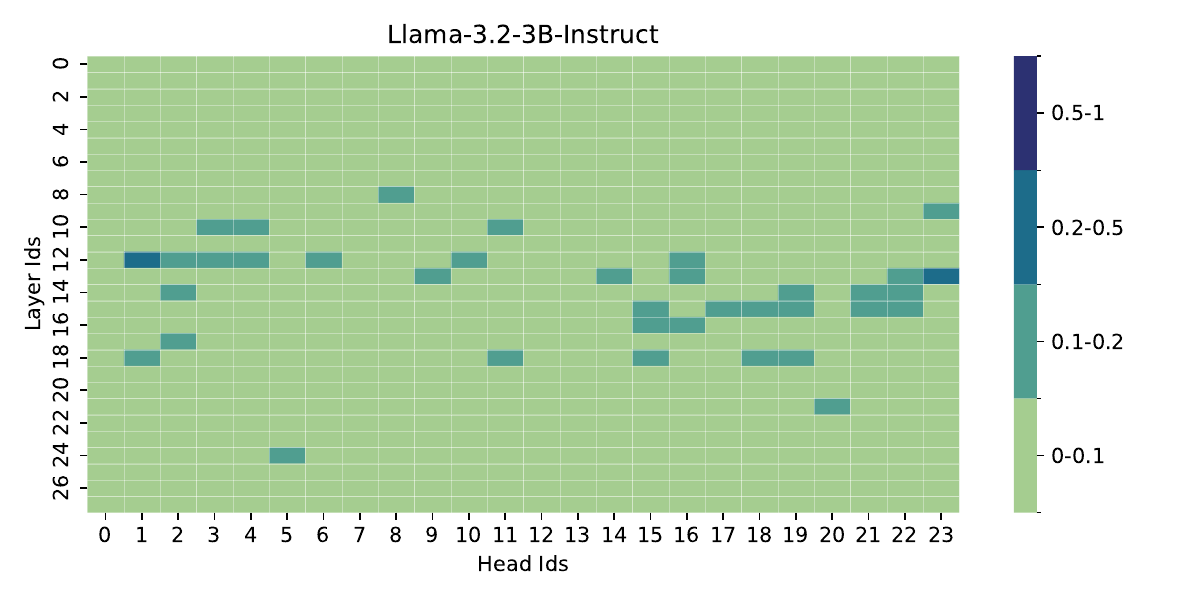}
\vspace{-4mm}
\caption{Location of the contextual heads.}
\label{fig:head_location}
\vspace{-4mm}
\end{minipage}
\end{wrapfigure}

\textbf{Contextual heads are sparse.} As shown in Figure \ref{fig:head_location}, among 672 attention heads in Llama-3.2-3B instruction model, only 2 (0.3\%) of the heads have a relevant contextual score that $>$ 0.2. Also, only 37 (5.5\%) of the heads have a relevant contextual score $>$0.1, and only 113 (16.8\%) of the heads have a relevant contextual score $>$0.05. In general, only a small amount of heads with high relevant contextual scores are considered to extract information from relevant contexts during autoregressive generation. Most heads, with low relevant contextual scores, are not considered to extract information from the relevant contexts.

\textbf{Contextual heads are mostly located in middle and late layers.} As shown in Figure \ref{fig:head_location}, most of the contextual heads with relevant contextual scores $>$0.1 are located from layer 8 to layer 18 (index from 0 to 27). 

\begin{table*}[]
\centering
\resizebox{0.97\textwidth}{!}{%
\begin{tabular}{@{}lllll|llll|llll|ll@{}}
\toprule
 & \multicolumn{4}{l}{Long} & \multicolumn{4}{l}{Correct} & \multicolumn{4}{l}{Wrong} & \multicolumn{2}{l}{Gold} \\ \midrule
Heads & R$\uparrow$ & IR$\downarrow$ & IR max$\downarrow$ & Sink & R$\uparrow$ & IR$\downarrow$ & IR max$\downarrow$ & Sink & R$\uparrow$ & IR$\downarrow$ & IR max$\downarrow$ & Sink & R & Sink \\
(13, 23) & 0.209 & 0.516 & 0.160 & 0.105 & 0.290 & 0.437 & 0.125 & 0.114 & 0.106 & 0.612 & 0.202 & 0.093 & 0.555 & 0.187 \\
(12, 1) & 0.203 & 0.568 & 0.161 & 0.079 & 0.283 & 0.490 & 0.129 & 0.084 & 0.106 & 0.664 & 0.201 & 0.070 & 0.637 & 0.153 \\
(15, 18) & 0.199 & 0.423 & 0.138 & 0.254 & 0.279 & 0.338 & 0.101 & 0.267 & 0.101 & 0.525 & 0.183 & 0.238 & 0.507 & 0.317 \\
(15, 22) & 0.195 & 0.391 & 0.140 & 0.244 & 0.277 & 0.309 & 0.103 & 0.254 & 0.098 & 0.487 & 0.184 & 0.227 & 0.481 & 0.280 \\
(14, 2) & 0.185 & 0.339 & 0.130 & 0.294 & 0.270 & 0.262 & 0.086 & 0.278 & 0.080 & 0.429 & 0.181 & 0.311 & 0.345 & 0.458 \\ \bottomrule
\end{tabular}%
}
\caption{Contextual scores of top-5 contextual heads. \textbf{Heads}: (Layer, head number), \textbf{R}: relevant contextual score, \textbf{IR}: irrelevant contextual score, \textbf{IR max}: max single document irrelevant contextual score, \textbf{Sink}: sink contextual score. We consider four cases: \textbf{Long}: standard 20-documents long context case. \textbf{Gold}: with only relevant contexts but not irrelevant ones. \textbf{Correct}: exactly matched for both gold and long case. \textbf{Wrong}: exactly matched for the gold case but not exactly matched in the long case. We define the correct and wrong based on the gold to filter out the cases that are not doable for LLMs.}
\label{tab:head_scores}
\end{table*}

\textbf{Contextual heads focus more on relevant context when the response is correct, focus less on relevant context when the response is wrong.}
As shown in Table \ref{tab:head_scores}, we found that overall, relevant contexts have lower scores than the irrelevant contexts since we have 19 documents as irrelevant context and only 1 as relevant context. However, in the long and correct case, the score for relevant context is larger than the IR max score. This means contextual heads have more focus on the relevant context when the generated answer is correct. While in the wrong case, this does not hold that relevant contexts have a lower score than the irrelevant ones with a max score.

\textbf{More attention is ``activated'' for long contexts compared to the short ones.}
As shown in Table \ref{tab:head_scores}, sink contextual scores are similar for long, correct, and wrong cases. However, the gold has a higher sink contextual score than the other three long context cases. At the same time, less attention is paid to the contexts for the gold cases than the three long context cases since the attentions are summed up to 1. This suggests that more attention is ``activated'' for long contexts compared to the short ones, and the sink contextual scores could be an indicator for such activation.

\subsection{Attention compensation on contextual heads}
\label{sec:attention_compensation}

From \S\ref{sec:properties_of_contextual_heads} we demonstrate the correct cases have a higher attention to the relevant contexts compared to the wrong cases. In this section, we aim to further demonstrate that if we could increase the attention to the relevant contexts for the contextual head, the distraction could be mitigated.

\begin{wrapfigure}{r}{0.66\textwidth}
\vspace{-4mm}
\centering
\begin{minipage}{0.66\textwidth}
\centering
\includegraphics[width=\linewidth]{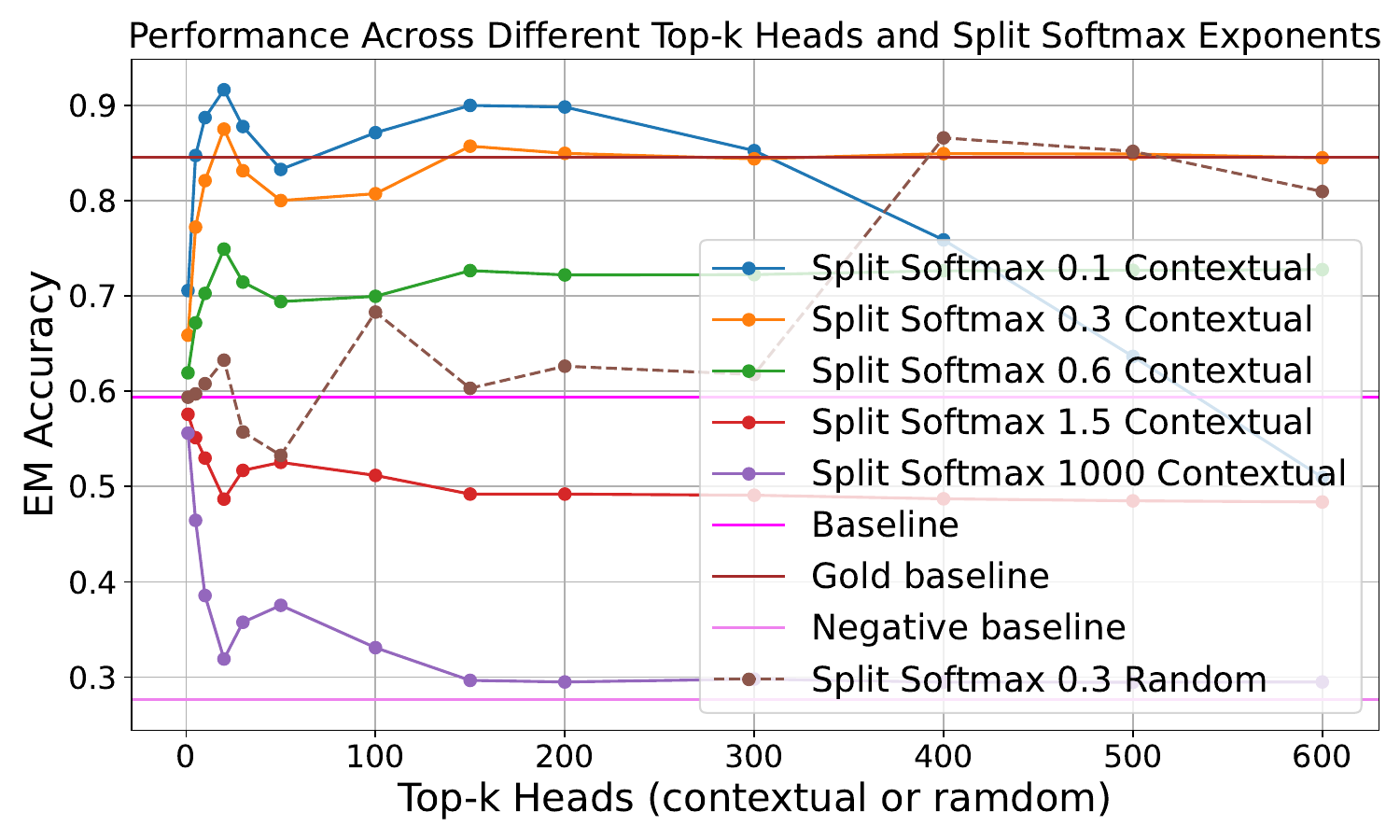}
\vspace{-4mm}
\caption{Performance across different top-$k$ contextual/random heads and split softmax exponents $\tau$. Baseline: 20 documents (1 relevant, 19 irrelevant) case without intervention. Gold baseline: 1 relevant document case without intervention. Negative baseline: 19 irrelevant documents case without intervention. }
\label{fig:attention_span}
\vspace{-4mm}
\end{minipage}
\end{wrapfigure}

\textbf{Attention compensation method.} We use split-softmax ~\cite{li2024measuring}, which can increase or decrease the attention on a token span for some specific attention heads. Specifically, given the attention weight matrix \( W \in \mathbb{R}^{T \times T} \) at layer \( \ell \) and head \( h \), we aim to modify the attention weights assigned to the relevant context span \( C = [c_{start}, \dots, c_{end}] \).
First, for each response token $r_i$ to be generated, we compute the total attention allocated to the span \( C \) by summing the relevant attention weights:
\begin{equation}
  \pi_C(i) = \sum_{j=c_{start}}^{c_{end}} W_{i,j}  
\end{equation}
We then rescale the attention distribution using the split-softmax transformation:
\begin{equation}
 W'_{i,j} =
\begin{cases} 
\frac{\pi_C(i)^\tau}{\pi_C(i)} W_{i,j}, & \text{if } j \in C \\
\frac{1-\pi_C(i)^\tau}{1-\pi_C(i)} W_{i,j}, & \text{if } j \notin C
\end{cases}   
\end{equation}
where \( \tau \) is the split softmax exponent controlling the strength of the modification, with \( \tau \geq 0 \). When \( 0 \leq \tau < 1 \), attention is increased for the span $C$, when \( \tau = 1 \), no modification is applied, and when $\tau>1$, attention is decreased for the span $C$. And smaller values of \( \tau \) increase the attention, while larger values of $\tau$ decrease the attention. The reweighted matrix \( W' \) ensures that the attention scores still sum to 1 across each row while redistributing more attention toward the span \( C \).

\textbf{Experiment settings.} 
We experiment with split softmax exponent $\tau = (0.1, 0.3, 0.6, 1.5, 1000)$ with the top-$k$ heads of $(1, 5, 10, 20, 30, 50, 100, 150, 200, 300, 400, 500, 600)$, using the testing split of our dataset. We also report the baseline EM accuracy of 0.59, which is without any split softmax intervention.

\textbf{Increasing attention to the relevant contexts mitigates the distraction, while decreasing attention to the relevant contexts results in more distraction.}
As shown in Figure \ref{fig:attention_span}, increasing the attention to the relevant contexts ($\tau < 1$) improves the performance. For all the cases of $\tau = (0.1, 0.3, 0.6)$, the EM accuracy is larger than the baseline. 
On the other hand, decreasing the attention to the relevant contexts ($\tau > 1$) decreases the performance. 

\textbf{Increasing attention on the contextual heads mitigates distraction, while increasing attention on non-contextual heads has a limited effect on distraction mitigation.}
We demonstrate this through two aspects: using top-$k$ contextual heads and $k$ random heads.
As shown in Figure \ref{fig:attention_span}, for the top-$k$ contextual heads, for all cases of $\tau < 1$, the EM accuracy improves with more attention heads being intervened from top-1 to top-20. The best EM accuracy (0.916) is achieved with top-20 heads and $\tau = 0.1$. However, with more top-$k$ heads intervened, the EM accuracy is decreased compared to the top-20 case. Notably, adding too much attention ($\tau = 0.1$) on 600 heads even makes the EM accuracy drop under the baseline. 
On the other hand, when using $k$ random heads with $\tau = 0.3$, we observe a limited ($<$0.3\%) EM accuracy improvement with $<$20 heads, a performance drop when using 50 heads, and a similar performance compared to contextual heads when using more than 400 heads. 
This demonstrates that increasing attention helps more with distraction mitigation when using contextual heads and helps less when using non-contextual heads.

\textbf{Contextual heads control the overall attention of the LLM.}
As shown in Figure \ref{fig:attention_span}, when intervening in top-20 contextual heads, increasing attention to the relevant context on the contextual heads, the EM accuracy can reach up to 0.916, better than the gold baseline of 0.847. On the other hand, with decreasing attention to the relevant context on the contextual heads, the EM accuracy can drop to 0.320, close to the negative baseline of 0.276. This suggests that the contextual heads control the overall attention of the LLM to the input tokens. In the case of increased attention on the contextual heads, the effect of input tokens in the relevant contexts can be amplified. In case of decreased attention on the contextual heads, the effect of input tokens in the relevant contexts can be nullified.

\section{Eliciting attention on relevant contexts via focus direction}
\label{sec:focus_direction}

From \S\ref{sec:attention_compensation} we show that increasing attention on the relevant contexts could mitigate the distraction. However, in practice, we do not have the label of relevant contexts during LLM inference. 
We wonder, can contextual heads figure out the relevant contexts by themselves?
Inspired by previous direction addition works ~\cite{turner2023activation, arditi2024refusal,li2024inference}, we hypothesize the existence of a focus direction that could make LLMs focus more on the relevant contexts. In this section, we first introduce a method to obtain the focus directions (\S\ref{sec:obtain_direction}). Then, we discuss the usage and effect of the focus directions (\S\ref{sec:apply_diection}).

\subsection{Obtaining focus direction }
\label{sec:obtain_direction}

To obtain the focus direction, we first need to identify the location of the focus direction. Previous works mainly focused on the residual stream activation ~\cite{turner2023activation, arditi2024refusal} or O projection~\cite{li2024inference} of attention heads, which do not have a direct relation with the attention and may not be feasible for our case. Since the attention is produced by key and query activation, we hypothesize that focus directions are situated within the key and query representation spaces. Based on the hypothesis, we aim to find two focus direction vectors, one for key activation and another for query activation for each attention head.

\textbf{Obtain focus directions by training.}
We consider a simple training method to obtain the focus direction. We first generate a response with $[I_p, C, I_q]$ (i.e., with relevant context only), obtain a gold LLM response $R_g$, for each sample in our training split, and obtain text sequences $[I_p, C, I_q, R_g]$.
We then cache the key activations $K \in \mathbb{R}^{T \times F}$ and query activations $Q \in \mathbb{R}^{T \times F}$ of the text sequence for each attention head, where $F$ is the feature dimension of $Q$ and $K$. The original attention weights is obtained by $W = \text{softmax}\left( \frac{QK^\top}{\sqrt{F}} \right)$. We add focus direction vectors $d_K \in \mathbb{R}^{F}$ and $d_Q \in \mathbb{R}^{F}$ for $K$ and $Q$, obtaining a new attention weights 
\begin{equation}
    W^d = \text{softmax}\left( \frac{(Q + d_Q)(K + d_K)^\top}{\sqrt{F}} \right)
\end{equation}
Given the new $W_d$, we can simply put it into Equation \ref{eq:relevant_score}, which obtains $S^d_C = \frac{1}{|R|} \sum_{i=r_{start}}^{r_{end}} {\sum_{j=c_{start}}^{c_{end}} {{W}^d_{i,j}}}$, measuring the attention to the relevant contexts $C$ when generating the LLM answer. We can use a simple loss function $L=-S^d_C$, training $d_K$ and $d_Q$ to obtain the focus direction. The directions maximize attention to the relevant contexts of the corresponding attention head during the response generation process.

\begin{wrapfigure}{r}{0.5\textwidth}
\vspace{-4mm}
\centering
\begin{minipage}{0.5\textwidth}
\centering
\includegraphics[width=\linewidth]{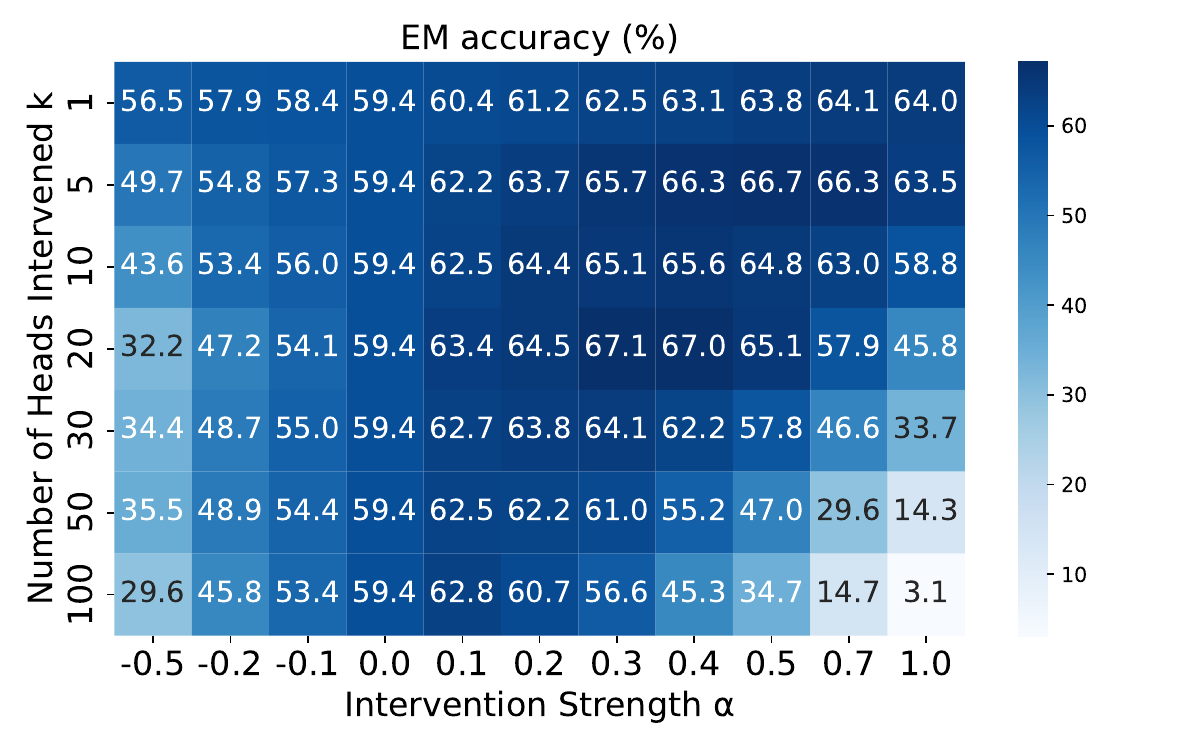}
\vspace{-4mm}
\caption{EM accuracy of different top-$k$ heads and $\alpha$.}
\label{fig:main}
\vspace{-4mm}
\end{minipage}
\end{wrapfigure}

\subsection{Inference time intervention with focus direction}
\label{sec:apply_diection}

Given focus direction $d_K$ and $d_Q$ for an attention head obtained by the previous step, we can apply them at the inference time with the following:
\begin{equation}
    W = \text{softmax}\left( \frac{(Q + \alpha d_Q)(K + \alpha d_K)^\top}{\sqrt{F}} \right)
\end{equation}
where $\alpha$ is an intervention factor to control the strength of the intervention. When $\alpha > 0$ is the positive intervention, aim to make the attention head pay more attention to the relevant context. When $\alpha < 0$ is the negative intervention, aim to make the attention head pay less attention to the relevant context. When $\alpha = 0$ no intervention is applied. In addition,  we can have a hyperparameter $k$ that intervenes top-$k$ contextual heads.

\subsection{Experiment settings}

We first cache the activations for the whole sequence of our training split and then obtain the focus directions by training. We used AdamW optimizer with a learning rate of $10^{-3}$ training for 10 epochs. For evaluation, we used our testing split. We report the contextual scores of the top-5 heads in Table \ref{tab:head_score_intervention} and the EM accuracy in Figure \ref{fig:main}.

\subsection{Results}
\label{sec:apply_diection_result}

\textbf{Focus directions make contextual heads pay more attention to the relevant context.}
As shown in Table \ref{tab:head_score_intervention}, when a positive focus direction is applied ($\alpha = 0.2$ and $\alpha = 0.5$), the contextual scores on the relevant context are increased. Also, the higher the $\alpha$, the more attention to the relevant contexts. On the other hand, when a negative focus direction is applied, the contextual scores on the relevant context are decreased.

\textbf{Focus direction kick the attention out of the sink.}
While increasing the attention to the relevant contexts, positive focus directions do not decrease the attention to the irrelevant contexts. Instead, the attention on irrelevant context may still have little increase. The main attention reassigned to the relevant contexts is from the attention sink. This suggests the main function of focus direction is to move the attention from the sink to the relevant contexts.

\textbf{Positive focus direction mitigates distraction, while negative focus direction leads to more distraction.}
As shown in Figure \ref{fig:main}, when applying a positive focus direction with $ 0 < \alpha \leq 0.5$, for the top 1-20 heads, the
EM accuracy has a consistent improvement compared to the baseline (59.4 \%). The best EM accuracy of 67.1\% was achieved with $\alpha = 0.3$ with top-20 heads\footnote{As noted in ~\cite{liu2024lost}, some distractor passages may contain a reasonable answer. As such, we don't expect the EM accuracy here to be comparable with the one in Figure \ref{fig:attention_span}.}. This demonstrates that positive focus directions could mitigate distraction. On the other hand, when applying a negative focus direction with $\alpha < 0$, the EM accuracy drops under the baseline, indicating more distraction than no intervention.

\textbf{Focus directions only help mitigate distraction on contextual heads.}
When applying a positive focus direction, we observe that an intervention of $> 20$ heads always results in lower EM accuracy than the one of 20 heads. This indicates focus direction only helps mitigate distraction on contextual heads. Applying focus direction on non-contextual heads may not help mitigate distraction. The observation is also consistent with the attention compensation result in Figure \ref{fig:attention_span}.

\textbf{Applying overly strong focus directions can inadvertently heighten attention to irrelevant contexts.}
As shown in Table \ref{tab:head_score_intervention}, from $\alpha=0.4$ to $\alpha =0.5$, the IR max score starts to rise at a higher rate than the R score. For example, for the head (13, 23), the R score increased from 0.40 to 0.41, and the IR max score increased from 0.21 to 0.23. The raised IR max score distracts the LLM, making the corresponding EM accuracy drop from 67.0\% to 65.1\%. Furthermore, when $\alpha=1.0$, the R score further drops to 0.34, and its value is similar to the IR max score. And the corresponding EM accuracy dropped 45.8\%, even worse than the baseline of 59.4. This indicates that applying a strong focus direction can also distract the LLM. The right level of focus is needed to align LLM to achieve optimal downstream performance.

\section{Focus directions are generalizable to different tasks}
\label{sec:focus_direction_tasks}

To study the effect of the focus direction on various long-context tasks, we use HELMET ~\cite{yen2024helmet}, a comprehensive benchmark for long-context evaluation. We use five categories of the task from HELMET, including 
\textbf{Synthetic recall (Recall)} (needle-in-a-haystack~\cite{hsieh2024ruler} and JSON KV retrieval task~\cite{liu2024lost}), 
\textbf{Retrieval-augmented generation (RAG)} (KILT benchmark ~\cite{petroni2020kilt}, including Natural Questions (NQ) ~\cite{kwiatkowski2019natural}, TriviaQA~\cite{joshi2017triviaqa}, HotpotQA~\cite{yang2018hotpotqa}, PopQA~\cite{mallen2022not}), 
\textbf{Passage re-ranking (Re-rank)} (MS MARCO ~\cite{bajaj2016ms}), 
\textbf{Many-shot in-context learning (ICL)} (TREC-course, TREC-fine ~\cite{li2002learning},  BANKING77 ~\cite{casanueva2020efficient}, CLINC150 ~\cite{larson2019evaluation}, NLU ~\cite{liu2021benchmarking}), \textbf{Long-document QA (Long QA)}(Infbench QA and multiple choice (MC) ~\cite{zhang2024infty}).

\begin{table}[]
\centering
\begin{minipage}{0.45\textwidth}
\centering
\resizebox{\textwidth}{!}{%
\begin{tabular}{lcccccc}
\hline
\multicolumn{7}{c}{ } \\
Model & Recall & RAG & Re-ranking & ICL & Long QA & Overall Average \\
\hline
\multicolumn{7}{l}{\textbf{Llama-3.2-3B}} \\
20\_-0.2 & \textcolor{red}{66.00} & \textcolor{red}{54.96} & \textcolor{green}{29.22} & \textcolor{black}{82.20} & - & \textcolor{red}{58.10} \\
10\_-0.2 & \textcolor{red}{73.81} & \textcolor{red}{56.58} & \textcolor{green}{26.73} & \textcolor{green}{83.00} & - & \textcolor{red}{60.03} \\
20\_0.2 & \textcolor{green}{81.50} & \textcolor{red}{58.75} & \textcolor{red}{25.37} & \textcolor{red}{80.20} & - & \textcolor{red}{61.46} \\
10\_0.2 & \textcolor{green}{82.00} & \textcolor{red}{58.54} & \textcolor{green}{26.16} & \textcolor{red}{80.60} & - & \textcolor{green}{61.83} \\
baseline & 78.88 & 58.83 & 26.10 & 82.20 & - & 61.50 \\
\multicolumn{7}{l}{\textbf{Llama-3.2-3B-Instruct}} \\
20\_-0.2 & \textcolor{red}{73.00} & \textcolor{red}{58.04} & \textcolor{red}{13.68} & \textcolor{red}{78.80} & \textcolor{green}{27.32} & \textcolor{red}{50.17} \\
10\_-0.2 & \textcolor{red}{79.12} & \textcolor{red}{60.21} & \textcolor{red}{13.32} & \textcolor{red}{79.40} & \textcolor{red}{26.66} & \textcolor{red}{51.74} \\
20\_0.2 & \textcolor{red}{83.50} & \textcolor{red}{60.25} & \textcolor{green}{20.58} & \textcolor{green}{80.60} & \textcolor{red}{26.09} & \textcolor{red}{54.20} \\
10\_0.2 & \textcolor{red}{83.69} & \textcolor{red}{62.08} & \textcolor{green}{20.77} & \textcolor{green}{80.40} & \textcolor{red}{25.94} & \textcolor{green}{54.58} \\
baseline & 84.38 & 63.00 & 17.13 & 80.20 & 26.78 & 54.30 \\
\multicolumn{7}{l}{\textbf{Qwen2.5-7B}} \\
20\_-0.2 & \textcolor{red}{94.56} & \textcolor{red}{53.50} & \textcolor{red}{22.86} & \textcolor{red}{77.60} & - & \textcolor{red}{62.13} \\
10\_-0.2 & \textcolor{red}{95.31} & \textcolor{green}{54.58} & \textcolor{green}{24.84} & \textcolor{red}{78.40} & - & \textcolor{green}{63.28} \\
20\_0.2 & \textcolor{red}{95.88} & \textcolor{red}{54.04} & \textcolor{green}{23.25} & \textcolor{red}{79.40} & - & \textcolor{red}{63.14} \\
10\_0.2 & \textcolor{red}{95.50} & \textcolor{red}{54.08} & \textcolor{red}{23.11} & \textcolor{green}{80.00} & - & \textcolor{red}{63.17} \\
baseline & 96.00 & 54.21 & 23.15 & 79.60 & - & 63.24 \\
\multicolumn{7}{l}{\textbf{Qwen2.5-7B-Instruct}} \\
20\_-0.2 & \textcolor{red}{94.38} & \textcolor{red}{55.87} & \textcolor{red}{35.88} & \textcolor{green}{78.40} & \textcolor{green}{33.73} & \textcolor{red}{59.65} \\
10\_-0.2 & \textcolor{green}{95.38} & \textcolor{red}{56.54} & \textcolor{red}{35.78} & \textcolor{green}{78.40} & \textcolor{green}{33.63} & \textcolor{green}{59.94} \\
20\_0.2 & \textcolor{green}{95.44} & \textcolor{green}{58.50} & \textcolor{green}{36.75} & \textcolor{green}{78.40} & \textcolor{green}{33.45} & \textcolor{green}{60.51} \\
10\_0.2 & \textcolor{green}{95.44} & \textcolor{green}{58.79} & \textcolor{red}{35.85} & \textcolor{green}{78.20} & \textcolor{green}{32.61} & \textcolor{green}{60.18} \\
baseline & 95.25 & 57.71 & 36.56 & 77.40 & 31.92 & 59.77 \\
\multicolumn{7}{l}{\textbf{Ministral-8B-Instruct-2410}} \\
20\_-0.2 & \textcolor{red}{94.62} & \textcolor{red}{61.79} & \textcolor{red}{31.31} & \textcolor{red}{77.20} & \textcolor{green}{33.59} & \textcolor{red}{59.70} \\
10\_-0.2 & \textcolor{red}{94.56} & \textcolor{red}{62.17} & \textcolor{red}{29.74} & \textcolor{red}{78.80} & \textcolor{green}{33.17} & \textcolor{red}{59.69} \\
20\_0.2 & \textcolor{red}{93.81} & \textcolor{red}{63.46} & \textcolor{green}{38.86} & \textcolor{green}{79.40} & \textcolor{red}{29.00} & \textcolor{green}{60.91} \\
10\_0.2 & \textcolor{red}{93.81} & \textcolor{green}{63.87} & \textcolor{green}{36.69} & \textcolor{green}{79.60} & \textcolor{red}{28.74} & \textcolor{green}{60.54} \\
baseline & 94.75 & 63.58 & 33.68 & 79.00 & 31.56 & 60.51 \\
\hline
\end{tabular}
}
\end{minipage}
\begin{minipage}{0.45\textwidth}
\centering
\resizebox{\textwidth}{!}{%
\begin{tabular}{lcccccc}
\hline
\multicolumn{7}{c}{ } \\
Model & Recall & RAG & Re-ranking & ICL & Long QA & Overall Average \\
\hline
\multicolumn{7}{l}{\textbf{Llama-3.2-3B}} \\
20\_-0.2 & \textcolor{red}{55.50} & \textcolor{red}{50.38} & \textcolor{red}{6.83} & \textcolor{red}{85.20} & - & \textcolor{red}{49.48} \\
10\_-0.2 & \textcolor{red}{64.56} & \textcolor{red}{53.96} & \textcolor{red}{6.24} & \textcolor{black}{86.20} & - & \textcolor{red}{52.74} \\
20\_0.2 & \textcolor{green}{66.31} & \textcolor{green}{56.46} & \textcolor{red}{7.08} & \textcolor{red}{85.40} & - & \textcolor{green}{53.81} \\
10\_0.2 & \textcolor{green}{65.69} & \textcolor{green}{55.83} & \textcolor{green}{9.27} & \textcolor{red}{85.40} & - & \textcolor{green}{54.05} \\
baseline & 65.50 & 54.83 & 7.29 & 86.20 & - & 53.46 \\
\multicolumn{7}{l}{\textbf{Llama-3.2-3B-Instruct}} \\
20\_-0.2 & \textcolor{red}{56.38} & \textcolor{red}{56.75} & \textcolor{green}{3.77} & \textcolor{red}{83.80} & \textcolor{red}{28.64} & \textcolor{red}{45.87} \\
10\_-0.2 & \textcolor{red}{61.06} & \textcolor{red}{58.33} & \textcolor{red}{2.44} & \textcolor{black}{85.00} & \textcolor{red}{30.38} & \textcolor{red}{47.44} \\
20\_0.2 & \textcolor{green}{65.81} & \textcolor{red}{59.21} & \textcolor{red}{2.72} & \textcolor{red}{83.40} & \textcolor{red}{28.23} & \textcolor{red}{47.87} \\
10\_0.2 & \textcolor{green}{64.25} & \textcolor{red}{59.79} & \textcolor{red}{3.10} & \textcolor{red}{84.20} & \textcolor{red}{26.80} & \textcolor{red}{47.63} \\
baseline & 64.12 & 59.96 & 3.77 & 85.00 & 31.13 & 48.80 \\
\multicolumn{7}{l}{\textbf{Qwen2.5-7B}} \\
20\_-0.2 & \textcolor{red}{42.06} & \textcolor{red}{41.96} & \textcolor{red}{1.30} & \textcolor{red}{77.40} & - & \textcolor{red}{40.68} \\
10\_-0.2 & \textcolor{red}{45.00} & \textcolor{red}{43.42} & \textcolor{green}{2.64} & \textcolor{red}{77.40} & - & \textcolor{red}{42.11} \\
20\_0.2 & \textcolor{green}{46.56} & \textcolor{red}{43.08} & \textcolor{red}{1.19} & \textcolor{red}{77.60} & - & \textcolor{red}{42.11} \\
10\_0.2 & \textcolor{green}{46.56} & \textcolor{red}{43.62} & \textcolor{red}{1.07} & \textcolor{green}{78.80} & - & \textcolor{green}{42.51} \\
baseline & 45.19 & 44.12 & 1.88 & 78.00 & - & 42.30 \\
\multicolumn{7}{l}{\textbf{Qwen2.5-7B-Instruct}} \\
20\_-0.2 & \textcolor{red}{46.31} & \textcolor{red}{43.96} & \textcolor{green}{11.92} & \textcolor{green}{78.80} & \textcolor{red}{22.07} & \textcolor{red}{40.61} \\
10\_-0.2 & \textcolor{red}{46.38} & \textcolor{red}{44.42} & \textcolor{green}{12.21} & \textcolor{black}{78.60} & \textcolor{red}{21.66} & \textcolor{red}{40.65} \\
20\_0.2 & \textcolor{green}{51.38} & \textcolor{green}{46.88} & \textcolor{red}{10.28} & \textcolor{red}{78.20} & \textcolor{green}{22.95} & \textcolor{green}{41.94} \\
10\_0.2 & \textcolor{green}{49.25} & \textcolor{green}{47.79} & \textcolor{red}{11.66} & \textcolor{red}{78.20} & \textcolor{green}{23.94} & \textcolor{green}{42.17} \\
baseline & 47.88 & 46.54 & 11.88 & 78.60 & 22.43 & 41.46 \\
\multicolumn{7}{l}{\textbf{Ministral-8B-Instruct-2410}} \\
20\_-0.2 & \textcolor{red}{30.56} & \textcolor{red}{46.17} & \textcolor{black}{0.00} & \textcolor{red}{80.60} & \textcolor{green}{21.41} & \textcolor{red}{35.75} \\
10\_-0.2 & \textcolor{red}{30.06} & \textcolor{red}{46.04} & \textcolor{black}{0.00} & \textcolor{red}{80.00} & \textcolor{red}{20.62} & \textcolor{red}{35.34} \\
20\_0.2 & \textcolor{green}{30.88} & \textcolor{red}{47.12} & \textcolor{black}{0.00} & \textcolor{green}{81.80} & \textcolor{red}{19.98} & \textcolor{red}{35.96} \\
10\_0.2 & \textcolor{green}{31.19} & \textcolor{red}{46.79} & \textcolor{black}{0.00} & \textcolor{green}{82.80} & \textcolor{red}{19.49} & \textcolor{red}{36.05} \\
baseline & 30.62 & 47.17 & 0.00 & 81.40 & 21.40 & 36.12 \\
\hline
\end{tabular}
}
\end{minipage}
\caption{Results of HELMET benchmark under 32k (left) and 64k (right) context. Green indicates better than the baseline; red indicates worse than the baseline.}
\label{tab:HELMET_main}
\end{table}

\textbf{Experiment settings.} We consider three LLMs, including Llama-3.2-3B-Instruct, Qwen2.5-7B-Instruct, and Ministral-8B-Instruct-2410. To show the effect of focus direction on base models, we also provide the results of Llama-3.2-3B and Qwen2.5-7B, using the focus direction obtained by their corresponding instruction models. We consider five settings, including baseline (no intervention), $\alpha = -0.2$, and $0.2$ for top-10 and top-20 attention heads.
Also, we experiment with 8k, 16k, 32k, 64k, and 128k token contexts, following the HELMET benchmark. We report the 32k and 64k results in Table \ref{tab:HELMET_main}, and the rest are in the tables in the appendix. We also report the sink contextual score under 8k and 16k contexts in Table \ref{tab:sink_8k} and \ref{tab:sink_16k}.

\textbf{Focus direction mitigates poor task alignment.}
We discuss this from two aspects. First, we compare the task performance between base models and instruction models. For a task, if there is a performance gain after post-training, the base model may have a performance gain by applying the focus direction. 
For example, for the HotpotQA task under 8k contexts (Llama), the performance improved from 52.67\% (base model) to 62.00\% after post-training. When focus directions are applied, the base model performance could be improved to 56.00\%. In this case, the base model does not have good task alignment and can benefit from applying focus direction.
Second, if there is an unusual sink contextual score, focus directions could help to achieve a better task alignment by paying the right amount of attention to the contexts. For example, for the TREC Coarse task under 8k contexts, the Llama-instruction model has a sink contextual score of 0.535, higher than the average score under 8k contexts of 0.297. As such, the LLM may not pay enough attention to the contexts. Adding a focus direction helps the performance improve from 69\% to 75\%.

\textbf{Focus directions help for the long context tasks that LLM could do well in the short context.}
For example, as shown in Table \ref{tab:HELMET_32k}, the Qwen instruction model has a high performance of 98\% on the MK Needle task under the 32k context. The performance drops to 48\% when it comes to the 64k contexts. Adding focus directions helps improve the performance to 63\%.

\textbf{Most of the tasks could be improved by either positive or negative focus direction.}
Table \ref{tab:HELMET_main} shows the category-based average performance of each task under 32k and 64k contexts. We found that 34 of the 48 task categories could have performance improvement by either positive or negative focus directions. This indicates that focus direction could play an important role in most of the long context tasks. This also confirms that the right level of focus is needed for optimal task performance. When an LLM exhibits excessive attention activation, a negative focus direction may help suppress irrelevant information. Conversely, when attention activation is insufficient, a positive focus direction can enhance attention to relevant contexts.

\textbf{Focus direction improves the overall performance of poorly aligned LLMs.}
We also show the overall average performance of all the tasks. We found that focus direction could improve the performance of 5 of 5 LLMs on 32k contexts and 3 of 5 LLMs on 64k contexts. We also check the standard deviation of the sink contextual scores of all the tasks for each LLM (Table \ref{tab:sink_8k} and \ref{tab:sink_16k}). We consider the LLMs with higher standard deviation poorly aligned since they do not have a consistent attention behavior under the same length of the contexts. 
Based on this, we consider the Qwen and LLama are more poorly aligned than Ministral. And over the performance of different tasks ranging from 8k to 128k contexts, Qwen and LLama have more improvement than the Ministral with the focus directions. 
We conclude that focus directions are likely to improve poorly aligned LLMs.

\section{Discussion}

\textbf{Contextual heads vs. retrieval heads.} 
A similar type of attention head with contextual heads is retrieval heads ~\cite{wu2024retrieval}. Retrieval heads are the attention heads used for copying tokens from the input to the output. 
We found that contextual heads are different from retrieval heads in the following aspects. 1) Location: As shown in Figure \ref{fig:contextual_vs_retrieval}, retrieval heads universally exist in different layers, while contextual heads are mainly located in the middle and late layers. Among the top 20 retrieval heads and contextual heads, only 5 of them overlap in the Llama-3.2-3B-Instruct model. 2) Function: retrieval heads focus on explicit copy tokens from the input to the output, while contextual heads control the overall attention of LLMs.

\textbf{Focus directions may be task dependent.}
While we verify the existence of the focus direction, we do not consider we locate ``optimal'' focus direction for every task. Instead, we consider the focus direction may be task-dependent. In other words, each task may have a different definition of relevant contexts and may have their corresponding focus directions. In addition, given optimal task focus directions, the overall level of attention activation may converge across tasks that share the same context length. We leave these as future work.

\textbf{Border impact of contextual heads and focus directions}
We consider that the focus direction may have the following applications: 
1) Focus directions may be an alternative approach for parameter-efficient fine-tuning ~\cite{xu2023parameter} for adapting long-context language models for different tasks. 
2) Focus directions may serve as a ``switch'' to control the LLM's use of contextual or internal knowledge, addressing knowledge conflicts ~\cite{xu2024knowledge}.

\section{Related work}

\textbf{Long context LLMs and evaluation.}
Advanced long-context LLMs now can accommodate 128k or more tokens in their context, including property models like GPT-4, Gemini, and Claude and open-source models like Llama 3.1 ~\cite{dubey2024llama}, 
Ministral and Qwen2.5 ~\cite{yang2024qwen2}. Such models enable various applications, such as long context QA ~\cite{wang2024novelqa, karpinska2024one}, in-context learning ~\cite{li2024long, agarwal2025many, bertsch2024context}, summarization ~\cite{chang2023booookscore,kim2024fables}, and retrieval-augmented generation ~\cite{lee2024can}. For evaluation, early works mainly focus on the synthetic tasks~\cite{hsieh2024ruler,liu2024lost, tay2020long}, such as the needle in the haystack, which may not well measure the LLM performance in the real world. Recent work has focused more on diverse and real-world settings, such as RAG~\cite{lee2024can}, in-context learning~\cite{li2024long}, and reasoning~\cite{zhou2025gsm}.

\textbf{Mechanistic interpretability on attention heads and activation steering.}
Our contextual heads relate to the recent work that discovers functional attention heads in LLMs, such as heads related to retrieval ~\cite{wu2024retrieval}, in-context learning ~\cite{olsson2022context, yin2025attention, ren2024identifying}, safety ~\cite{chen2024finding}, and knowledge conflicts ~\cite{jin2024cutting, shi2024ircan}. Our focus direction is related to the activation steering work, which could use a directional vector to control the LLMs' behavior, such as truthfulness ~\cite{li2024inference}, sentiment ~\cite{han2023word}, and refusal ~\cite{arditi2024refusal}.

\section{Conclusion}

In this paper, we identify the contextual heads, which control the overall attention of LLMs, and focus directions on the contextual heads that could make LLMs pay more attention to the relevant contexts. We first propose a contextual scoring method to identify the contextual heads. Then, we demonstrate that insufficient attention to the relevant context in these heads is the cause of LLM distraction. Moreover, we identify focus directions, which could move the attention of contextual heads from the attention sink to the relevant contexts and thus mitigate the distraction. 
We further study the effect of focus directions on the real-world long context benchmark and find that focus directions could help mitigate poor task alignment. At last, we discuss the potential border impact of focus directions for long-context LLM alignment.



\bibliography{custom}
\bibliographystyle{colm2025_conference}

\appendix
\section{Appendix}

\subsection{Details of Experiment on HELMET benchmark}

We used the same settings as the HELMET benchmark. For metrics, we used the substring exact match for all the retrieval-augmented
generation and synthetic recall tasks, NDCG@10 for the passage re-ranking task, and accuracy for all many-shot in-context learning tasks, ROUGE F1 for Infbench QA, and accuracy for the Infbench MC. We exclude other tasks that require model-based evaluation.

\begin{figure}[h]
    \centerline{\includegraphics[width=0.60\textwidth]{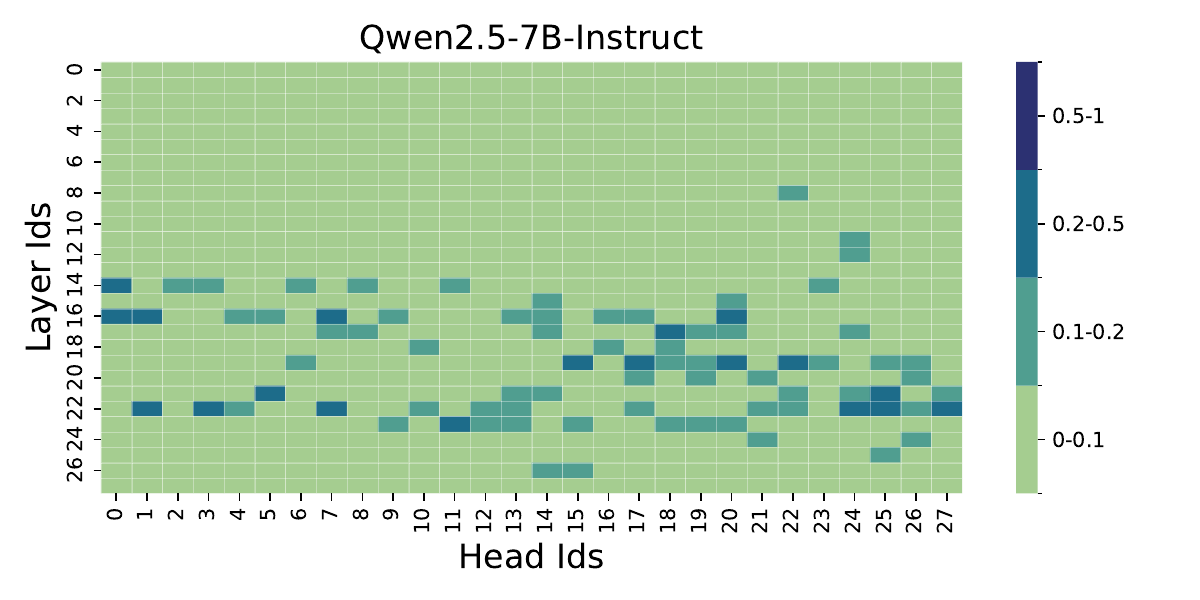}}
\caption{Location of the contextual heads of Qwen2.5-7B-Instruct.}
\label{fig:head_location_qwen}
\end{figure}

\begin{figure}[h]
    \centerline{\includegraphics[width=0.60\textwidth]{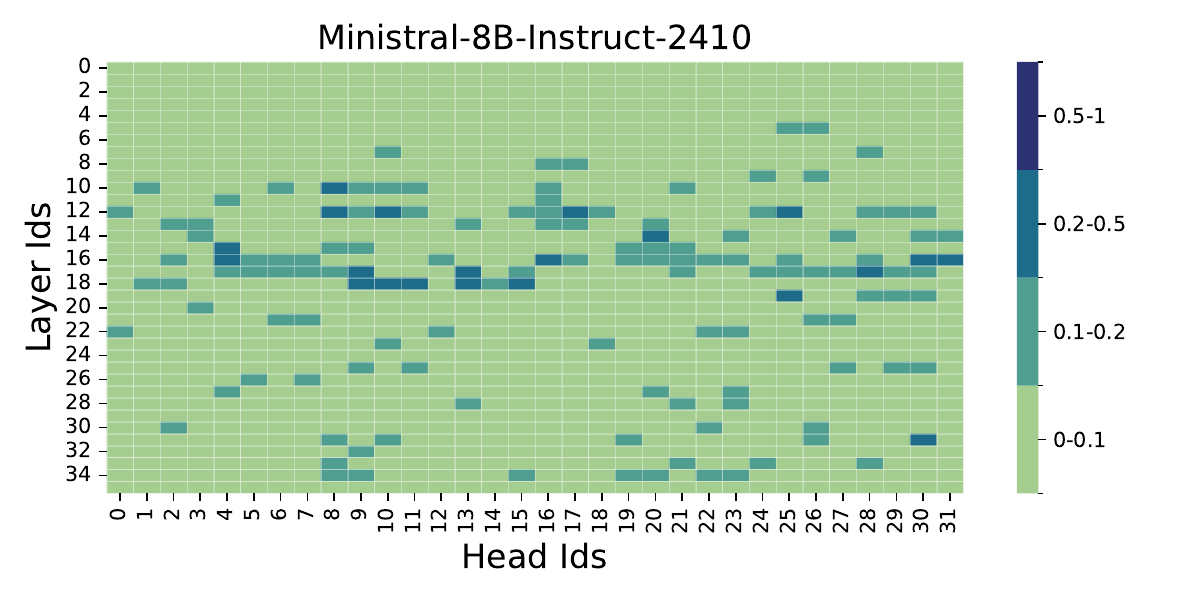}}
\caption{Location of the contextual heads Ministral-8B-Instruct-2410.}
\label{fig:head_location_ministral}
\end{figure}

\begin{figure*}[h]
  \centering
  \includegraphics[width=0.60\textwidth]{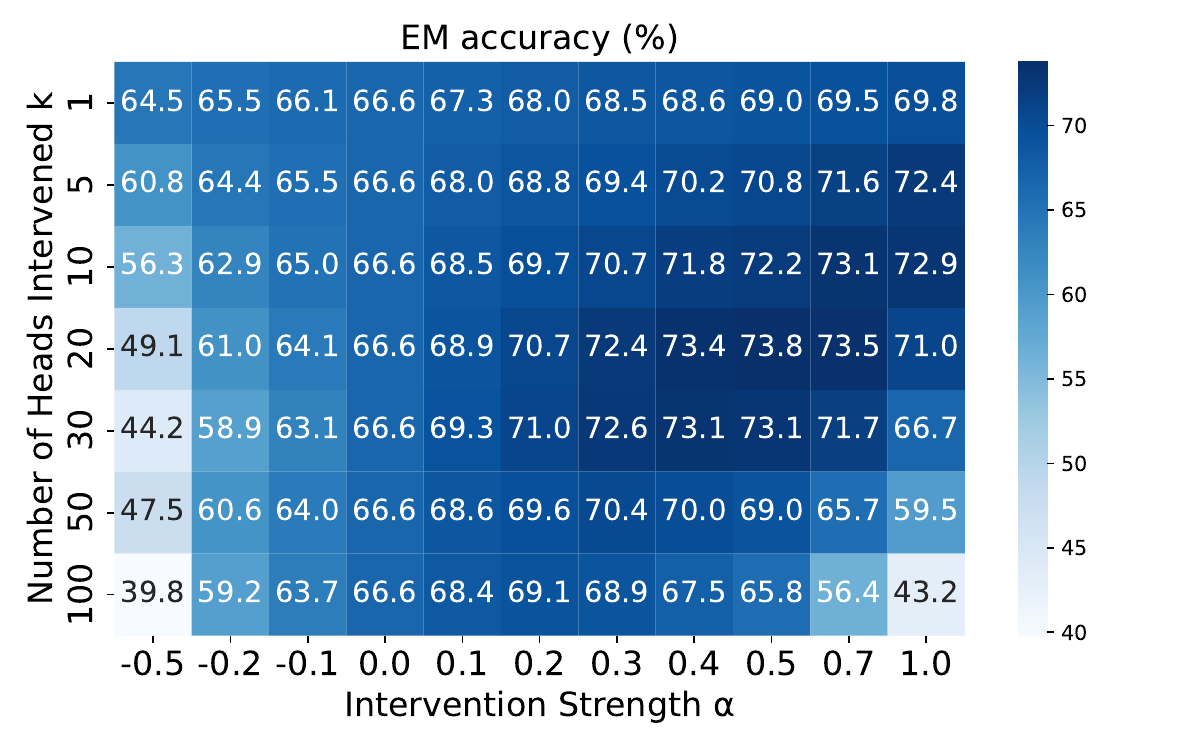}
  \caption{EM accuracy of different top-$k$ heads and $\alpha$ of Qwen2.5-7B-Instruct.}
  \label{fig:main_qwen}
\end{figure*}

\begin{figure*}[h]
  \centering
  \includegraphics[width=0.60\textwidth]{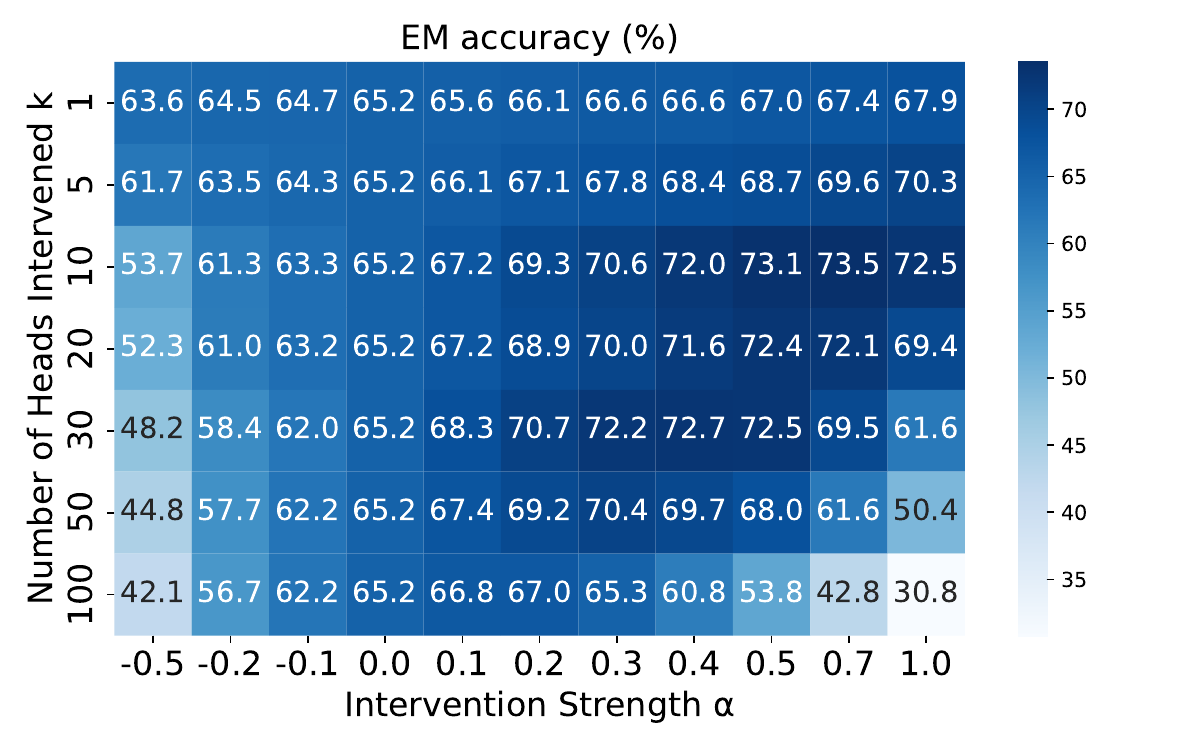}
  \caption{EM accuracy of different top-$k$ heads and $\alpha$ of Ministral-8B-Instruct-2410.}
  \label{fig:main_mistral}
\end{figure*}

\begin{figure*}[h]
  \centering
  \includegraphics[width=0.70\textwidth]{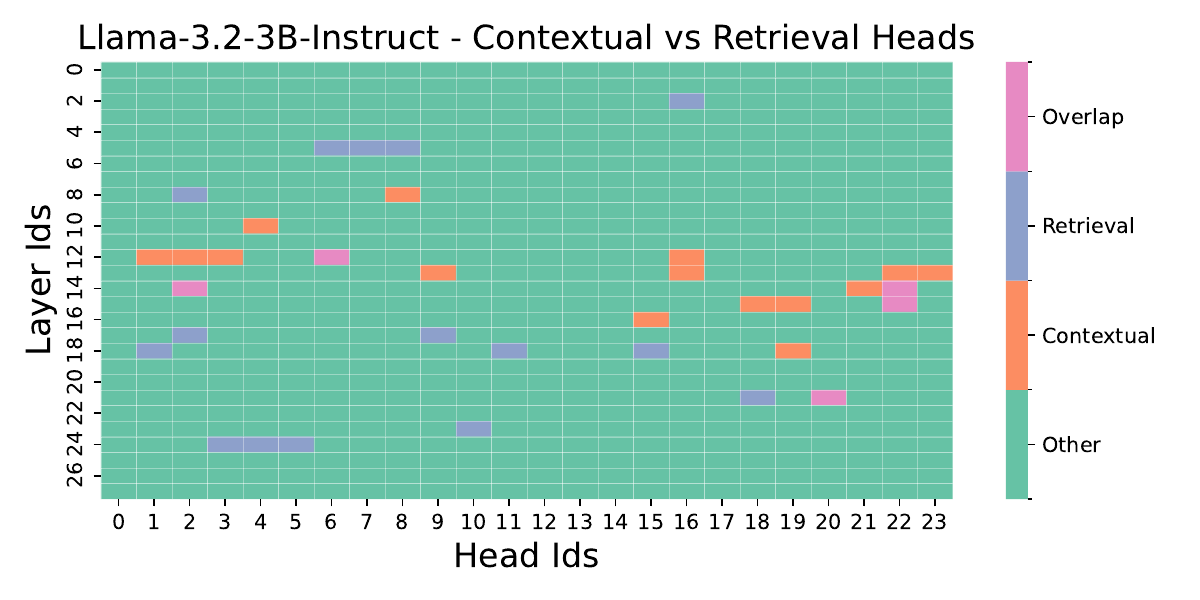}
  \caption{The location of top-20 contextual head vs. top-20 retrieval heads.}
  \label{fig:contextual_vs_retrieval}
\end{figure*}

\begin{table}[h]
\centering
\begin{minipage}{0.61\textwidth}
\centering
\resizebox{0.98\textwidth}{!}{%
\begin{tabular}{@{}lllllll@{}}
\toprule
Heads    & R            & IR           & IR max       & Sink         & R Gold       & Sink Gold    \\ \midrule
\multicolumn{7}{l}{\textbf{$\alpha = -0.2$}}                                                                  \\
(13, 23) & 0.10 (-0.11) & 0.43 (-0.08) & 0.12 (-0.04) & 0.30 (+0.19) & 0.30 (-0.26) & 0.42 (+0.23) \\
(12, 1)  & 0.12 (-0.08) & 0.56 (-0.01) & 0.14 (-0.02) & 0.14 (+0.06) & 0.45 (-0.19) & 0.31 (+0.15) \\
(15, 18) & 0.09 (-0.11) & 0.35 (-0.08) & 0.10 (-0.04) & 0.44 (+0.18) & 0.25 (-0.26) & 0.55 (+0.24) \\
(15, 22) & 0.09 (-0.10) & 0.31 (-0.08) & 0.10 (-0.04) & 0.40 (+0.16) & 0.26 (-0.23) & 0.49 (+0.21) \\
(14, 2)  & 0.07 (-0.12) & 0.21 (-0.13) & 0.08 (-0.05) & 0.56 (+0.27) & 0.12 (-0.23) & 0.72 (+0.26) \\ \hline
\multicolumn{7}{l}{\textbf{$\alpha = 0.2$}}                                                                   \\
\bottomrule
\end{tabular}%
}
\end{minipage}
\begin{minipage}{0.30\textwidth}
\centering
\resizebox{0.98\textwidth}{!}{%
\begin{tabular}{@{}lllll@{}}
\toprule
Heads & R & IR & IR max & Sink \\ \midrule
\multicolumn{4}{l}{\textbf{$\alpha = 0.4$}}    \\
(13, 23) & 0.40784 & 0.46287 & 0.21073 & 0.00626 \\
(12, 1) & 0.40324 & 0.50762 & 0.21712 & 0.01094 \\
(15, 18) & 0.37799 & 0.48941 & 0.20475 & 0.04639 \\
(15, 22) & 0.36006 & 0.48624 & 0.20482 & 0.04563 \\
(14, 2) & 0.44528 & 0.45260 & 0.23089 & 0.01582 \\ \midrule
\multicolumn{5}{l}{\textbf{$\alpha = 0.5$}}   \\
(13, 23) & 0.41196 & 0.47068 & 0.23630 & 0.00324 \\
(12, 1) & 0.41427 & 0.51336 & 0.24182 & 0.00608 \\
(15, 18) & 0.38408 & 0.51542 & 0.22953 & 0.02423 \\
(15, 22) & 0.36751 & 0.52062 & 0.23100 & 0.02447 \\
(14, 2) & 0.45509 & 0.47693 & 0.26239 & 0.00799 \\ \midrule
\multicolumn{5}{l}{\textbf{$\alpha = 1.0$}} \\
(13, 23) & 0.34991 & 0.55895 & 0.34943 & 0.00008 \\
(12, 1) & 0.36854 & 0.59642 & 0.33347 & 0.00010 \\
(15, 18) & 0.33626 & 0.61708 & 0.32412 & 0.00017 \\
(15, 22) & 0.31526 & 0.64934 & 0.32020 & 0.00054 \\
(14, 2) & 0.38478 & 0.58332 & 0.36628 & 0.00018 \\ \bottomrule
\end{tabular}%
}
\end{minipage}
\caption{\textbf{Left:} Contextual scores of top-5 contextual heads when top-5 heads are intervened. The value in the ``()" represents the difference compared to the result without intervention in Table \ref{tab:head_scores}. \textbf{Right:} Contextual scores of top-5 contextual heads when top-20 heads are intervened.}
\label{tab:head_score_intervention}
\end{table}

\begin{table}[h]
\centering
\resizebox{\textwidth}{!}{%
\begin{tabular}{lcccccccccccccccc}
\hline
\multicolumn{1}{c}{} & \multicolumn{4}{c}{Recall} & \multicolumn{4}{c}{RAG} & \multicolumn{1}{c}{Re-ranking} & \multicolumn{5}{c}{ICL} & \multicolumn{2}{c}{Long QA} \\
model & MK Needle & MK UUID & MV & JSON KV & NQ & TriviaQA & HotpotQA & PopQA & MS MARCO & TREC Coarse & TREC Fine & BANKING77 & CLINC150 & NLU & Infbench QA & Infbench MC \\
\hline
\multicolumn{17}{l}{\textbf{Llama-3.2-3B}} \\
20\_-0.2 & \textcolor{red}{90.00} & \textcolor{red}{89.00} & \textcolor{red}{53.00} & \textcolor{red}{98.00} & \textcolor{red}{48.67} & \textcolor{green}{82.33} & \textcolor{red}{48.33} & \textcolor{red}{52.67} & \textcolor{red}{38.15} & \textcolor{red}{58.00} & \textcolor{red}{27.00} & \textcolor{red}{66.00} & \textcolor{red}{73.00} & \textcolor{red}{77.00} & - & - \\
10\_-0.2 & \textcolor{black}{99.00} & \textcolor{red}{95.00} & \textcolor{red}{60.50} & \textcolor{green}{100.00} & \textcolor{red}{50.67} & \textcolor{green}{82.33} & \textcolor{red}{50.67} & \textcolor{red}{54.00} & \textcolor{red}{42.31} & \textcolor{red}{59.00} & \textcolor{red}{28.00} & \textcolor{red}{66.00} & \textcolor{red}{75.00} & \textcolor{red}{78.00} & - & - \\
20\_0.2 & \textcolor{red}{98.00} & \textcolor{red}{97.00} & \textcolor{red}{64.50} & \textcolor{red}{98.00} & \textcolor{black}{54.50} & \textcolor{red}{79.33} & \textcolor{green}{55.67} & \textcolor{red}{56.17} & \textcolor{green}{46.21} & \textcolor{green}{67.00} & \textcolor{green}{34.00} & \textcolor{red}{68.00} & \textcolor{green}{78.00} & \textcolor{black}{82.00} & - & - \\
10\_0.2 & \textcolor{black}{99.00} & \textcolor{black}{99.00} & \textcolor{green}{67.25} & \textcolor{red}{98.00} & \textcolor{green}{54.67} & \textcolor{red}{81.00} & \textcolor{green}{56.00} & \textcolor{green}{56.83} & \textcolor{red}{42.41} & \textcolor{green}{67.00} & \textcolor{green}{35.00} & \textcolor{black}{69.00} & \textcolor{green}{78.00} & \textcolor{red}{81.00} & - & - \\
\hline
\end{tabular}
}
\caption{Results of HELMET benchmark under 8k context.}
\label{tab:HELMET_8k}
\end{table}

\begin{table}[h]
\centering
\resizebox{\textwidth}{!}{%
\begin{tabular}{lcccccccccccccccc}
\hline
\multicolumn{1}{c}{} & \multicolumn{4}{c}{Recall} & \multicolumn{4}{c}{RAG} & \multicolumn{1}{c}{Re-ranking} & \multicolumn{5}{c}{ICL} & \multicolumn{2}{c}{Long QA} \\
model & MK Needle & MK UUID & MV & JSON KV & NQ & TriviaQA & HotpotQA & PopQA & MS MARCO & TREC Coarse & TREC Fine & BANKING77 & CLINC150 & NLU & Infbench QA & Infbench MC \\
\hline
\multicolumn{17}{l}{\textbf{Llama-3.2-3B}} \\
20\_-0.2 & \textcolor{red}{94.00} & \textcolor{red}{73.00} & \textcolor{red}{66.75} & \textcolor{red}{95.00} & \textcolor{red}{47.00} & \textcolor{green}{86.33} & \textcolor{red}{43.67} & \textcolor{red}{48.00} & \textcolor{red}{30.15} & \textcolor{black}{82.00} & \textcolor{red}{44.00} & \textcolor{red}{83.00} & \textcolor{red}{87.00} & \textcolor{red}{80.00} & - & - \\
10\_-0.2 & \textcolor{black}{99.00} & \textcolor{red}{84.00} & \textcolor{red}{73.75} & \textcolor{black}{96.00} & \textcolor{red}{50.83} & \textcolor{green}{87.00} & \textcolor{red}{46.33} & \textcolor{red}{50.50} & \textcolor{red}{32.46} & \textcolor{black}{82.00} & \textcolor{red}{42.00} & \textcolor{red}{83.00} & \textcolor{green}{90.00} & \textcolor{red}{79.00} & - & - \\
20\_0.2 & \textcolor{red}{98.00} & \textcolor{green}{91.00} & \textcolor{red}{66.25} & \textcolor{black}{96.00} & \textcolor{red}{52.50} & \textcolor{red}{82.50} & \textcolor{green}{51.33} & \textcolor{green}{54.83} & \textcolor{red}{31.44} & \textcolor{green}{83.00} & \textcolor{red}{42.00} & \textcolor{red}{87.00} & \textcolor{green}{90.00} & \textcolor{red}{81.00} & - & - \\
10\_0.2 & \textcolor{black}{99.00} & \textcolor{black}{90.00} & \textcolor{green}{74.25} & \textcolor{black}{96.00} & \textcolor{red}{53.17} & \textcolor{red}{84.17} & \textcolor{black}{50.00} & \textcolor{green}{54.17} & \textcolor{red}{31.54} & \textcolor{green}{84.00} & \textcolor{red}{42.00} & \textcolor{black}{89.00} & \textcolor{black}{89.00} & \textcolor{green}{83.00} & - & - \\
\hline
\end{tabular}
}
\caption{Results of HELMET benchmark under 16k context.}
\label{tab:HELMET_16k}
\end{table}

\begin{table}[h]
\centering
\centering
\resizebox{\textwidth}{!}{%
\begin{tabular}{lcccccccccccccccc}
\hline
\multicolumn{1}{c}{} & \multicolumn{4}{c}{Recall} & \multicolumn{4}{c}{RAG} & \multicolumn{1}{c}{Re-ranking} & \multicolumn{5}{c}{ICL} & \multicolumn{2}{c}{Long QA} \\
model & MK Needle & MK UUID & MV & JSON KV & NQ & TriviaQA & HotpotQA & PopQA & MS MARCO & TREC Coarse & TREC Fine & BANKING77 & CLINC150 & NLU & Infbench QA & Infbench MC \\
\hline
\multicolumn{17}{l}{\textbf{Llama-3.2-3B}} \\
20\_-0.2 & \textcolor{red}{88.00} & \textcolor{red}{43.00} & \textcolor{red}{54.00} & \textcolor{red}{79.00} & \textcolor{red}{45.33} & \textcolor{red}{84.67} & \textcolor{red}{43.00} & \textcolor{red}{46.83} & \textcolor{green}{29.22} & \textcolor{red}{87.00} & \textcolor{green}{59.00} & \textcolor{black}{90.00} & \textcolor{red}{89.00} & \textcolor{green}{86.00} & - & - \\
10\_-0.2 & \textcolor{red}{95.00} & \textcolor{red}{54.00} & \textcolor{red}{56.25} & \textcolor{black}{90.00} & \textcolor{red}{44.17} & \textcolor{red}{86.83} & \textcolor{red}{46.00} & \textcolor{red}{49.33} & \textcolor{green}{26.73} & \textcolor{red}{86.00} & \textcolor{green}{62.00} & \textcolor{green}{92.00} & \textcolor{red}{89.00} & \textcolor{green}{86.00} & - & - \\
20\_0.2 & \textcolor{green}{98.00} & \textcolor{green}{77.00} & \textcolor{red}{60.00} & \textcolor{green}{91.00} & \textcolor{red}{47.50} & \textcolor{red}{86.17} & \textcolor{green}{49.67} & \textcolor{green}{51.67} & \textcolor{red}{25.37} & \textcolor{green}{89.00} & \textcolor{red}{53.00} & \textcolor{red}{89.00} & \textcolor{red}{87.00} & \textcolor{red}{83.00} & - & - \\
10\_0.2 & \textcolor{green}{98.00} & \textcolor{green}{75.00} & \textcolor{red}{59.00} & \textcolor{green}{96.00} & \textcolor{red}{46.67} & \textcolor{green}{87.67} & \textcolor{green}{49.33} & \textcolor{red}{50.50} & \textcolor{green}{26.16} & \textcolor{green}{89.00} & \textcolor{red}{53.00} & \textcolor{red}{88.00} & \textcolor{red}{89.00} & \textcolor{red}{84.00} & - & - \\
\hline
\end{tabular}
}
\caption{Results of HELMET benchmark under 32k context.}
\label{tab:HELMET_32k}
\end{table}

\begin{table}[h]
\centering
\resizebox{\textwidth}{!}{%
\begin{tabular}{lcccccccccccccccc}
\hline
\multicolumn{1}{c}{} & \multicolumn{4}{c}{Recall} & \multicolumn{4}{c}{RAG} & \multicolumn{1}{c}{Re-ranking} & \multicolumn{5}{c}{ICL} & \multicolumn{2}{c}{Long QA} \\
model & MK Needle & MK UUID & MV & JSON KV & NQ & TriviaQA & HotpotQA & PopQA & MS MARCO & TREC Coarse & TREC Fine & BANKING77 & CLINC150 & NLU & Infbench QA & Infbench MC \\
\hline
\multicolumn{17}{l}{\textbf{Llama-3.2-3B}} \\
20\_-0.2 & \textcolor{red}{86.00} & \textcolor{red}{27.00} & \textcolor{red}{50.00} & \textcolor{red}{59.00} & \textcolor{red}{39.17} & \textcolor{red}{82.83} & \textcolor{red}{40.00} & \textcolor{red}{39.50} & \textcolor{red}{6.83} & \textcolor{red}{89.00} & \textcolor{red}{68.00} & \textcolor{red}{91.00} & \textcolor{black}{92.00} & \textcolor{green}{86.00} & - & - \\
10\_-0.2 & \textcolor{green}{93.00} & \textcolor{red}{37.00} & \textcolor{green}{60.25} & \textcolor{red}{68.00} & \textcolor{red}{40.33} & \textcolor{green}{87.67} & \textcolor{green}{45.00} & \textcolor{red}{42.83} & \textcolor{red}{6.24} & \textcolor{red}{89.00} & \textcolor{black}{72.00} & \textcolor{red}{91.00} & \textcolor{black}{92.00} & \textcolor{green}{87.00} & - & - \\
20\_0.2 & \textcolor{red}{85.00} & \textcolor{green}{55.00} & \textcolor{red}{54.25} & \textcolor{green}{71.00} & \textcolor{black}{45.33} & \textcolor{red}{84.17} & \textcolor{green}{47.33} & \textcolor{green}{49.00} & \textcolor{red}{7.08} & \textcolor{black}{90.00} & \textcolor{red}{70.00} & \textcolor{red}{91.00} & \textcolor{green}{93.00} & \textcolor{red}{83.00} & - & - \\
10\_0.2 & \textcolor{red}{87.00} & \textcolor{green}{51.00} & \textcolor{red}{54.75} & \textcolor{black}{70.00} & \textcolor{red}{44.67} & \textcolor{red}{83.83} & \textcolor{green}{45.33} & \textcolor{green}{49.50} & \textcolor{green}{9.27} & \textcolor{black}{90.00} & \textcolor{red}{69.00} & \textcolor{black}{92.00} & \textcolor{black}{92.00} & \textcolor{red}{84.00} & - & - \\
\hline
\end{tabular}
}
\caption{Results of HELMET benchmark under 64k context.}
\label{tab:HELMET_64k}
\end{table}

\begin{table}[h]
\centering
\resizebox{\textwidth}{!}{%
\begin{tabular}{lcccccccccccccccc}
\hline
\multicolumn{1}{c}{} & \multicolumn{4}{c}{Recall} & \multicolumn{4}{c}{RAG} & \multicolumn{1}{c}{Re-ranking} & \multicolumn{5}{c}{ICL} & \multicolumn{2}{c}{Long QA} \\
model & MK Needle & MK UUID & MV & JSON KV & NQ & TriviaQA & HotpotQA & PopQA & MS MARCO & TREC Coarse & TREC Fine & BANKING77 & CLINC150 & NLU & Infbench QA & Infbench MC \\
\hline
\multicolumn{17}{l}{\textbf{Llama-3.2-3B}} \\
20\_-0.2 & \textcolor{red}{72.00} & \textcolor{red}{13.00} & \textcolor{red}{47.00} & \textcolor{red}{57.00} & \textcolor{red}{35.83} & \textcolor{red}{77.17} & \textcolor{red}{37.33} & \textcolor{red}{40.67} & \textcolor{red}{4.74} & \textcolor{red}{89.00} & \textcolor{red}{77.00} & \textcolor{red}{92.00} & \textcolor{red}{92.00} & \textcolor{red}{88.00} & - & - \\
10\_-0.2 & \textcolor{red}{76.00} & \textcolor{red}{20.00} & \textcolor{red}{51.75} & \textcolor{red}{59.00} & \textcolor{red}{35.33} & \textcolor{red}{80.83} & \textcolor{red}{41.67} & \textcolor{green}{42.67} & \textcolor{red}{4.62} & \textcolor{black}{90.00} & \textcolor{black}{81.00} & \textcolor{red}{92.00} & \textcolor{red}{91.00} & \textcolor{red}{87.00} & - & - \\
20\_0.2 & \textcolor{red}{84.00} & \textcolor{red}{23.00} & \textcolor{green}{61.50} & \textcolor{red}{60.00} & \textcolor{green}{38.33} & \textcolor{red}{81.33} & \textcolor{red}{43.00} & \textcolor{red}{42.00} & \textcolor{green}{6.78} & \textcolor{red}{89.00} & \textcolor{red}{76.00} & \textcolor{red}{92.00} & \textcolor{black}{93.00} & \textcolor{black}{89.00} & - & - \\
10\_0.2 & \textcolor{green}{87.00} & \textcolor{red}{25.00} & \textcolor{red}{53.75} & \textcolor{red}{62.00} & \textcolor{red}{37.00} & \textcolor{red}{81.67} & \textcolor{green}{44.00} & \textcolor{green}{43.00} & \textcolor{green}{6.56} & \textcolor{red}{89.00} & \textcolor{red}{76.00} & \textcolor{black}{93.00} & \textcolor{black}{93.00} & \textcolor{green}{90.00} & - & - \\
\hline
\end{tabular}
}
\caption{Results of HELMET benchmark under 128k context.}
\label{tab:HELMET_128k}
\end{table}

\begin{table}[h]
\centering
\resizebox{0.60\textwidth}{!}{%
\begin{tabular}{lcccccc}
\hline
\multicolumn{7}{c}{ } \\
Model & Recall & RAG & Re-ranking & ICL & Long QA & Overall Average \\
\hline
\multicolumn{7}{l}{\textbf{Llama-3.2-3B}} \\
20\_-0.2 & \textcolor{red}{82.50} & \textcolor{red}{58.00} & \textcolor{red}{38.15} & \textcolor{red}{60.20} & - & \textcolor{red}{59.71} \\
10\_-0.2 & \textcolor{red}{88.62} & \textcolor{red}{59.42} & \textcolor{red}{42.31} & \textcolor{red}{61.20} & - & \textcolor{red}{62.89} \\
20\_0.2 & \textcolor{red}{89.38} & \textcolor{red}{61.42} & \textcolor{green}{46.21} & \textcolor{green}{65.80} & - & \textcolor{green}{65.70} \\
10\_0.2 & \textcolor{green}{90.81} & \textcolor{green}{62.12} & \textcolor{red}{42.41} & \textcolor{green}{66.00} & - & \textcolor{green}{65.34} \\
baseline & 90.62 & 61.46 & 43.52 & 64.40 & - & 65.00 \\
\hline
\end{tabular}
}
\caption{Category average results of HELMET benchmark under 8k context.}
\label{tab:HELMET_8k_avg}
\end{table}

\begin{table}[h]
\centering
\resizebox{0.60\textwidth}{!}{%
\begin{tabular}{lcccccc}
\hline
\multicolumn{7}{c}{ } \\
Model & Recall & RAG & Re-ranking & ICL & Long QA & Overall Average \\
\hline
\multicolumn{7}{l}{\textbf{Llama-3.2-3B}} \\
20\_-0.2 & \textcolor{red}{82.19} & \textcolor{red}{56.25} & \textcolor{red}{30.15} & \textcolor{red}{75.20} & - & \textcolor{red}{60.95} \\
10\_-0.2 & \textcolor{red}{88.19} & \textcolor{red}{58.67} & \textcolor{red}{32.46} & \textcolor{red}{75.20} & - & \textcolor{red}{63.63} \\
20\_0.2 & \textcolor{red}{87.81} & \textcolor{red}{60.29} & \textcolor{red}{31.44} & \textcolor{red}{76.60} & - & \textcolor{red}{64.04} \\
10\_0.2 & \textcolor{green}{89.81} & \textcolor{red}{60.38} & \textcolor{red}{31.54} & \textcolor{black}{77.40} & - & \textcolor{red}{64.78} \\
baseline & 89.75 & 60.92 & 32.71 & 77.40 & - & 65.20 \\
\hline
\end{tabular}
}
\caption{Category average results of HELMET benchmark under 16k context.}
\label{tab:HELMET_16k_avg}
\end{table}

\begin{table}[h]
\centering
\resizebox{0.60\textwidth}{!}{%
\begin{tabular}{lcccccc}
\hline
\multicolumn{7}{c}{ } \\
Model & Recall & RAG & Re-ranking & ICL & Long QA & Overall Average \\
\hline
\multicolumn{7}{l}{\textbf{Llama-3.2-3B}} \\
20\_-0.2 & \textcolor{red}{47.25} & \textcolor{red}{47.75} & \textcolor{red}{4.74} & \textcolor{red}{87.60} & - & \textcolor{red}{46.83} \\
10\_-0.2 & \textcolor{red}{51.69} & \textcolor{red}{50.12} & \textcolor{red}{4.62} & \textcolor{red}{88.20} & - & \textcolor{red}{48.66} \\
20\_0.2 & \textcolor{red}{57.12} & \textcolor{red}{51.17} & \textcolor{green}{6.78} & \textcolor{red}{87.80} & - & \textcolor{red}{50.72} \\
10\_0.2 & \textcolor{red}{56.94} & \textcolor{green}{51.42} & \textcolor{green}{6.56} & \textcolor{red}{88.20} & - & \textcolor{red}{50.78} \\
baseline & 58.50 & 51.17 & 5.53 & 89.20 & - & 51.10 \\
\hline
\end{tabular}
}
\caption{Category average results of HELMET benchmark under 128k context.}
\label{tab:HELMET_128k_avg}
\end{table}

\begin{table}[h]
\centering
\resizebox{\textwidth}{!}{%
\begin{tabular}{@{}llllllllllllllllll@{}}
\toprule
model & MK Needle & MK UUID & MV & JSON KV & NQ & TriviaQA & HotpotQA & PopQA & MS MARCO & TREC Coarse & TREC Fine & BANKING77 & CLINC150 & NLU & Infbench QA & Infbench MC & STD \\ \midrule
Llama-3.2-3B & 21.60 & 23.61 & 26.68 & 22.63 & 19.63 & 21.51 & 19.82 & 21.83 & 22.43 & 29.28 & 23.04 & 7.78 & 19.70 & 22.22 & - & - & 7.21 \\
Llama-3.2-3B-Instruct & 32.53 & 23.09 & 23.01 & 29.37 & 22.28 & 25.65 & 24.15 & 24.00 & 32.52 & 44.59 & 36.98 & 22.30 & 24.11 & 25.65 & 17.69 & 31.37 & 9.33 \\
Qwen2.5-7B & 13.61 & 5.25 & 6.44 & 6.43 & 8.68 & 9.71 & 7.74 & 9.15 & 4.53 & 30.56 & 6.50 & 2.80 & 4.49 & 7.73 & - & - & 6.94 \\
Qwen\_Qwen2.5-7B-Instruct & 9.45 & 5.00 & 8.04 & 4.89 & 7.55 & 10.15 & 8.57 & 9.23 & 5.98 & 19.65 & 6.96 & 2.45 & 4.81 & 6.96 & 0.25 & 0.43 & 4.70 \\
Ministral-8B-Instruct-2410 & 19.87 & 12.69 & 12.41 & - & - & 13.26 & 13.02 & 16.48 & 10.44 & 16.29 & 12.64 & 6.81 & 11.28 & 13.35 & 11.02 & 15.21 & 4.55 \\ \bottomrule
\end{tabular}%
}
\caption{Sink contextual scores (\%) and its standard deviation (STD) under 8k contexts (average of top-5 contextual heads).}
\label{tab:sink_8k}
\end{table}

\begin{table}[h]
\centering
\resizebox{\textwidth}{!}{%
\begin{tabular}{@{}llllllllllllllllll@{}}
\toprule
model & MK Needle & MK UUID & MV & JSON KV & NQ & TriviaQA & HotpotQA & PopQA & MS MARCO & TREC Coarse & TREC Fine & BANKING77 & CLINC150 & NLU & Infbench QA & Infbench MC & STD \\ \midrule
Llama-3.2-3B & 21.91 & 25.35 & 23.94 & 23.19 & 20.44 & 21.88 & 20.91 & 23.70 & 23.74 & 29.52 & 30.02 & 9.78 & 26.57 & 23.73 & - & - & 4.81 \\
Llama-3.2-3B-Instruct & 31.85 & 29.34 & 24.42 & 34.02 & 22.53 & 26.78 & 24.81 & 25.90 & 30.04 & 53.52 & 35.39 & 26.45 & 31.76 & 31.67 & 18.61 & 28.49 & 7.71 \\
Qwen2.5-7B & 16.80 & 6.39 & 8.35 & 6.61 & 9.25 & 10.63 & 8.82 & 10.03 & 5.35 & 31.29 & 20.61 & 3.40 & 11.72 & 10.63 & - & - & 7.24 \\
Qwen2.5-7B-Instruct & 20.07 & 18.41 & 10.42 & 6.87 & 6.97 & 8.89 & 9.10 & 10.69 & 4.18 & 33.32 & 23.57 & 3.58 & 9.67 & 7.86 & 0.17 & 0.28 & 8.87 \\
Ministral-8B-Instruct-2410 & 22.45 & 15.48 & 14.76 & 14.56 & 11.59 & 16.52 & 15.89 & 20.63 & 12.32 & 26.98 & 12.92 & 9.70 & 16.21 & 14.71 & 17.16 & 20.32 & 4.39 \\ \bottomrule
\end{tabular}%
}
\caption{Sink contextual scores (\%) and its standard deviation (STD) under 16k contexts (average of top-5 contextual heads).}
\label{tab:sink_16k}
\end{table}

\end{document}